\documentclass[letterpaper, 10 pt, conference]{ieeeconf}  
\IEEEoverridecommandlockouts                              
\usepackage[english]{babel}
\usepackage[T1]{fontenc}

\usepackage[nolist]{acronym}
\usepackage[cmex10]{amsmath}
\usepackage[font={small}]{caption}
\usepackage{subcaption}
\usepackage{amssymb}
\usepackage{bm}
\usepackage{booktabs}
\usepackage{color}
\usepackage{float}
\usepackage{graphicx}
\usepackage{hyperref}
\usepackage{import}
\usepackage{mathtools}
\usepackage{multirow}
\usepackage{multicol}
\usepackage{flushend}
\usepackage{outline}
\usepackage{paralist}
\usepackage{siunitx}
\usepackage{stmaryrd}
\usepackage{systeme}
\usepackage{url}
\usepackage{tikz}
\usepackage{booktabs}
\usepackage{balance}
\usepackage{cite}
\usetikzlibrary{tikzmark}
\usepackage{esvect}
\usepackage[normalem]{ulem}
\usepackage{bbm}
\usepackage{siunitx}
\usepackage{wrapfig,lipsum,booktabs}

\usepackage{caption}
\usepackage{subcaption}
\usepackage{graphicx}
\usepackage{subcaption}
\usepackage{adjustbox}
\usepackage{xcolor}
\usepackage{flushend}
\usepackage{booktabs}
\usepackage{threeparttable}

\usepackage{colortbl}
\definecolor{ourcolor}{HTML}{99e0eb}
\definecolor{ourblue}{HTML}{27a2c3}
\definecolor{tablecolor}{HTML}{ccf2f5} 
\definecolor{tablecolor2}{HTML}{ffcdb4}
\definecolor{citecolor}{HTML}{fe7b5b}
\definecolor{grey}{rgb}{0.9, 0.9, 0.9}

\definecolor{gred}{rgb}{0.859,0.267,0.216}
\definecolor{ggreen}{rgb}{0.059,0.616,0.345}
\definecolor{deepblue}{HTML}{27a2c3}
\definecolor{deepred}{HTML}{fe7b5b}

\definecolor{orange}{RGB}{255, 128, 0}
\definecolor{blue}{RGB}{0, 0, 255}
\definecolor{red}{RGB}{255, 0, 0}
\definecolor{green}{RGB}{0, 255, 0}

\title{\LARGE \bf Dynamic Legged Ball Manipulation on Rugged Terrains with \\Hierarchical Reinforcement Learning}
\author{Dongjie Zhu$^{1}$, Zhuo Yang$^{2}$, Tianhang Wu$^{2}$, Luzhou Ge$^{2}$, Xuesong Li$^{2}$, Qi Liu$^{3}$, Xiang Li$^{1}$
\thanks{
$^{1}$Department of Automation, Tsinghua University.
}
\thanks{$^{2}$School of Computer Science, Beijing Institute of Technology.}
\thanks{$^{3}$Faculty of Robot Science and Engineering, Northeastern University.}
\thanks{
Corresponding authors: Qi Liu, Xiang Li (liuqi@mail.neu.edu.cn, xiangli@tsinghua.edu.cn).}
\thanks{This work was supported in part by the Science and Technology Innovation 2030-Key Project under Grant 2021ZD0201404, in part by the National Natural Science Foundation of China under Grant U21A20517 and 62461160307, and in part by the BNRist project under Grant BNR2024TD03003.}
}

\begin{document}
\maketitle

\begin{abstract}
Advancing the dynamic loco-manipulation capabilities of quadruped robots in complex terrains is crucial for performing diverse tasks. 
Specifically, dynamic ball manipulation in rugged environments presents two key challenges. The first is coordinating distinct motion modalities to integrate terrain traversal and ball control seamlessly. The second is overcoming sparse rewards in end-to-end deep reinforcement learning, which impedes efficient policy convergence.
To address these challenges, we propose a hierarchical reinforcement learning framework. A high-level policy, informed by proprioceptive data and ball position, adaptively switches between pre-trained low-level skills such as ball dribbling and rough terrain navigation.
We further propose Dynamic Skill-Focused Policy Optimization to suppress gradients from inactive skills and enhance critical skill learning.
Both simulation and real-world experiments validate that our methods outperform baseline approaches in dynamic ball manipulation across rugged terrains, highlighting its effectiveness in challenging environments.
Videos are on our website: \url{dribble-hrl.github.io}.

\end{abstract}

\section{Introduction}
With rapid advancements in robotics and artificial intelligence, quadruped robots can perform agile movements such as running\cite{margolis2024rapid}, parkour\cite{zhuang2023robot}, and football shooting\cite{ji2022hierarchical}, showing great potential in various practical applications. 
Currently, loco-manipulation—manipulating objects while executing dynamic movements—still remains a major research challenge\cite{ha2024learning}.
By reusing the legs for manipulation, loco-manipulation enhances the multi-tasking ability of quadruped robots and reduces operational costs. 
This capability is crucial for tasks such as search-and-rescue, package delivery, and robot soccer competitions.
Ball manipulation, as a canonical benchmark for such loco-manipulation tasks, has received extensive research attention in recent years.

As a dynamic mobile manipulation task, ball manipulation necessitates a seamless integration of visual perception, dynamic locomotion, and object manipulation.
DribbleBot\cite{Ji2023DribbleBot} undergoes extensive training in the simulator using deep reinforcement learning (RL) and domain randomization, demonstrating its ability to perform zero-shot transfer to dribble the ball on flat ground.
DexDribbler\cite{hu2024dexdribbler} builds upon the previous work by introducing a feedback control reward term and a context-aided estimator to improve the robot's ball control capabilities.
However, these works lack the capability to traverse complex terrains and manipulate dynamic objects on rugged surfaces.

\begin{figure}[t]
    \centering
    \includegraphics[width=1.0\linewidth]{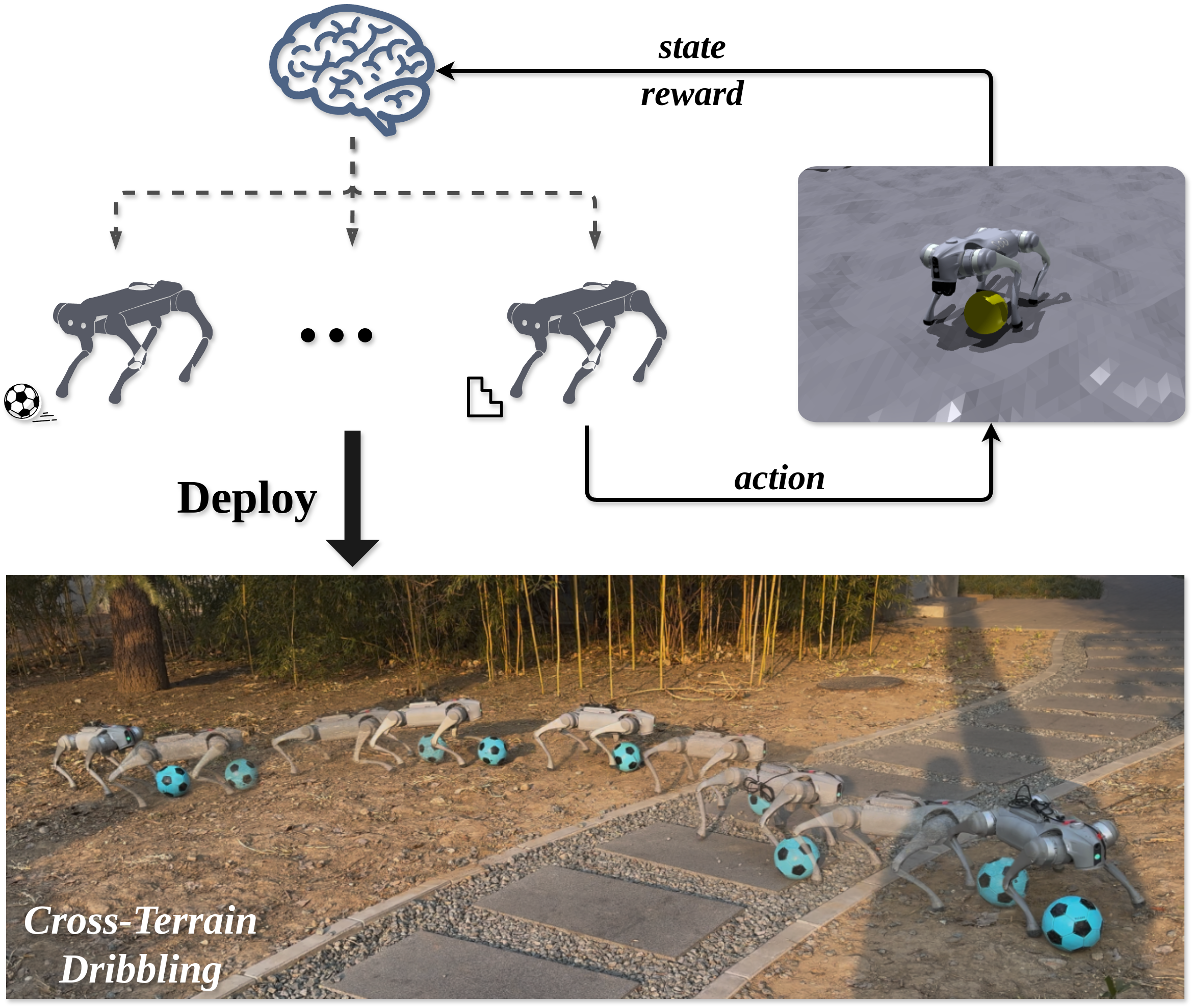}
    \caption{\textbf{Demonstration of legged ball dribbling with hierarchical framework.}
    The high-level policy selects and coordinates pre-trained dribbling and locomotion skills for dynamic ball manipulation, optimized through deep RL.
    We deploy the trained policy in the real world via zero-shot transfer, enabling the robot to perform cross-terrain dribbling.
    }
    \label{fig:headfig}
    \vspace{-0.3cm}
\end{figure}

This paper aims to design an efficient learning framework for legged ball manipulation on rugged terrains.
The main challenge is that ball manipulation and locomotion on rugged terrains are two distinct motion modalities that cannot be addressed by the same control strategy.
Additionally, end-to-end deep RL approach\cite{Ji2023DribbleBot, hu2024dexdribbler} struggles to provide useful reward signals for the initial policy on such terrain, thereby inhibiting exploration.
To solve this problem, we adopt hierarchical RL, integrating policies for both ball dribbling and terrain traversal, as presented in Fig. \ref{fig:headfig}.
The high-level policy perceives the terrain and ball position in real time, enabling agile switching between low-level dribbling and locomotion skills to maintain ball control.
To address learning efficiency and convergence challenges in mixed discrete-continuous action spaces for this task, we propose a dynamic skill-focused loss formulation by suppressing gradients from inactive skills and amplifying critical ones, improving convergence and stability.
Furthermore, reward design and curriculum learning are used to improve training, boosting policy convergence and task completion.

The main contributions of our work are as follows:
\begin{itemize}
    \item A hierarchical RL framework for agile control over low-level skills to achieve dynamic ball manipulation on rugged terrains.
    \item A novel loss formulation, Dynamic Skill-Focused Policy Optimization (DSF-PO), designed for handling the simultaneous discrete and continuous outputs of the high-level policy. 
    \item Demonstration of ball manipulation on rugged terrains in simulation and successful transfer to a real-world robot.
\end{itemize}

This research enables quadruped robots to perform dynamic ball manipulation on complex terrains, with potential applications in emergency supply transport in disaster zones, military operations, and robot soccer competitions.

\section{Related Work}
\subsection{Locomotion and Manipulation with Quadruped Robots}
Recent advances in learning-based methods have enabled quadruped robots to perform complex locomotion and manipulation tasks\cite{nahrendra2023dreamwaq, ji2022concurrent, margolis2024rapid, kim2024not}.
A noticeable trend is that the controller of quadruped robots would be trained in the physics simulator through large-scale GPU parallelism and then deployed in the real world\cite{hwangbo2019learning, lee2020learning, rudin2022learning, miki2022learning, margolis2023walk}.
The teacher-student paradigm \cite{lee2020learning, kumar2021rma} and learning across various posture configurations \cite{margolis2023walk} have contributed to robust locomotion on challenging terrains. Vision-based deep RL \cite{zhuang2023robot} has also been used to navigate complex environments.

Existing learning-based methods typically enable quadruped robots to manipulate objects using their front limbs by either generating end-effector (toe) trajectories or directly optimizing policies through reward functions.
Notable approaches include the RL-based end-effector tracker \cite{arm2024pedipulate}, trajectory interpolation \cite{cheng2023legs} and RL-based motion tracking networks \cite{ji2022hierarchical, huang2023creating, he2024learning, huang2024hilma}, which follow Bezier curves generated by the upper-level planner to guide the toe's interaction with the ball.
More complex tasks like ball dribbling in motion are addressed through deep RL with domain randomization \cite{Ji2023DribbleBot, hu2024dexdribbler}.

However, existing works have relatively limited performance and hence cannot deal with more complex tasks such as legged ball manipulation on rugged terrains.
This motivates our approach to seamlessly integrate locomotion and dribbling strategies within a unified framework.

\subsection{Hierarchical RL for Robot Control}
Hierarchical RL frameworks, such as the options framework \cite{precup2000temporal} and goal-conditioned hierarchies \cite{nachum2018nearoptimal}, have proven effective in robot control.
The \emph{compositionality} arising from the core mechanisms of temporal and state abstractions in hierarchical RL is regarded as a key factor contributing to its advantages in credit assignment, exploration, and continual learning\cite{make4010009}.
Specifically, high-level actions are executed at a lower temporal frequency than the atomic actions of the environment, effectively reducing episode length\cite{strehl2009reinforcement, azar2017minimax}. These high-level actions often correspond to behaviors with greater semantic meaning, making exploration and learning more efficient\cite{nachum2019does}.

Hierarchical frameworks in robot control are primarily classified into two categories:
\begin{itemize}
    \item  Utilizing a task-independent lower-level controller and a task-specific upper-level controller.
    The lower-level controller can function as either a learning-based trajectory tracker\cite{peng2017deeploco, jain2019hierarchical, huang2024hilma, he2024learning} or a model-based trajectory generator\cite{tan2023hierarchical}.
    Depending on its design, the upper-level controller may provide the lower-level controller with either goal positions\cite{gehring2021hierarchical, yang2021hierarchical} or trajectory parameters\cite{ji2022hierarchical, huang2023creating}.
    \item Integrating multiple skills. 
    \cite{kumar2023cascaded, yokoyama2023asc} incorporate diverse skills using a residual module to accomplish more complex tasks. This approach enables the integration of multiple low-level skills, each operating with distinct observation spaces, action spaces, or modalities.
\end{itemize}

Inspired by hierarchical control frameworks that leverage both task decomposition and multi-skill integration, we adopt this approach for legged ball dribbling. 
By leveraging a hierarchical structure, our method ensures flexible skill coordination and enhances adaptability to diverse environments.

\section{Hierarchical Policy Architecture}
\begin{figure*}[h]
    \centering
    \includegraphics[width=1.0\textwidth]{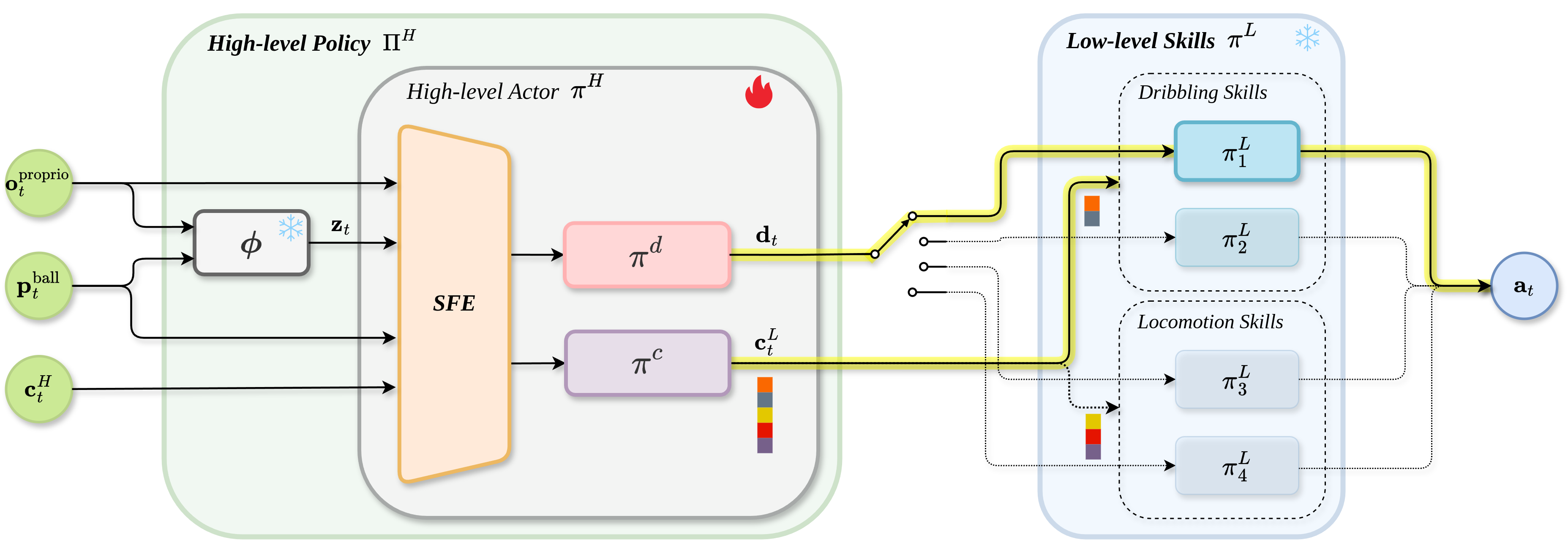}
    \caption{\textbf{Proposed hierarchical framework. }
    The figure illustrates that when the high-level actor outputs $\mathbf{d}_t=1$, only $\pi_1^L$ is activated, with the first two dimensions of $\mathbf{c}_t^L$ provided as input.
    The context-aided estimator network $\boldsymbol{\phi}$ and all low-level skills $\pi^L$ are frozen during training.}
    \label{fig:architecture}
    \vspace{-0.3cm}
\end{figure*}

As depicted in Fig. \ref{fig:architecture}, we propose a novel hierarchical framework that empowers a quadruped robot to execute dynamic and agile ball manipulation on rugged terrains.
This section introduces the low-level policy, which comprises dribbling and locomotion skills, and details the high-level policy architecture along with its training environment.
Sec. \ref{sec:DSF-PO} presents DSF-PO, a customized loss formulation for PPO tailored to this task.
Sec. \ref{sec:training strategy} outlines the training strategy for the high-level policy, covering reward design and curriculum learning.

\subsection{Learning Low-level Skills}\label{subsec:low-level skills}
For dribbling tasks over complex terrains, neither locomotion nor loco-manipulation alone performs well. Instead, the problem can be framed as a long-horizon task composed of locomotion and loco-manipulation sub-tasks.
The loco-manipulation sub-task involves dribbling when near the ball, while the locomotion sub-task focuses on approaching the ball when far away or navigating rough terrain. 
To address this, we employ four low-level policies: two for dribbling and two for locomotion, each tailored to specific scenarios. These policies, trained end-to-end using deep RL, function as skill modules that can be flexibly combined for more complex tasks and real-world deployment.

\vspace{0.13em}
\subsubsection{Dribbling Skills}
The dribbling skills aim to enable the quadruped robot to move the ball in a given direction and speed through front limb manipulation. Building on the models proposed in\cite{Ji2023DribbleBot, hu2024dexdribbler} , we train two distinct dribbling policies $\pi^L_1$ and $\pi^L_2$. These policies apply kicks with varying amplitudes, allowing the robot to dribble effectively across diverse terrains.

\subsubsection{Locomotion Skills}
Cross-terrain loco-manipulation places significant demands on the quadruped robot’s locomotion capabilities, requiring enhanced adaptability to sloped and rough terrains—conditions that cannot be adequately trained during ball dribbling.
To address this, we incorporate locomotion skills inspired by \cite{margolis2023walk, chen2024slr}, defining $\pi^L_3$ for fast movement on relatively flat terrain and $\pi^L_4$ for robust traversal over rough terrain.

\vspace{0.5em}
The observation space for each skill comprises the union of low-level commands $\mathbf{c}_t^L$, a subset of proprioceptive data $\mathbf{p}^{\text{proprio}}$, timing reference variables $\boldsymbol{\theta}_t$ and other task-specific information. The low-level commands $\mathbf{c}^L_t$ can represent either the target ball velocity command ($\mathbf{v}^\text{cmd}_x$ and $\mathbf{v}^\text{cmd}_y$) or the locomotion command ($\mathbf{v}^\text{cmd}_x$, $\mathbf{v}^\text{cmd}_y$, and $\boldsymbol{\omega}$). For dribbling skills, ball position $\mathbf{p}^\text{ball}$ is also required.
Each skill generates target joint positions $\mathbf{a}_t \in \mathbb{R}^{12}$ at 50 Hz, which serve as inputs to a PD controller for driving joint motors. All low-level skills are trained end-to-end using Proximal Policy Optimization (PPO) \cite{schulman2017proximal} within the Isaac Gym \cite{rudin2022learning}.

\subsection{High-level Policy}\label{subsec:high-level policy}
While low-level skills handle joint control for dribbling and locomotion, the high-level policy ensures seamless transitions between these motion modalities. Different from \cite{cheng2023legs}, we leverage deep RL for efficient and adaptable high-level policy training.
The whole high-level policy neural network $\Pi^H$ consists of two blocks: a context-aided estimator network $\boldsymbol{\phi}$ and an actor $\pi^H$, as presented in Fig. \ref{fig:architecture}.
Inspired by \cite{hu2024dexdribbler}, we build the context-aided estimator $\boldsymbol{\phi}$ to estimate terrain parameters, the quadruped robot's posture, and the ball's motion state in an explicit way. The estimator is trained under supervised learning using ground-truth states provided by the simulator.

\subsubsection{Observation Space}
The high-level observation space $\mathbf{O}^H$ is composed of three key components: proprioceptive data, ball position, and user commands. 
The context-aided estimator $\boldsymbol{\phi}$ processes the first two components, while the actor $\pi^H$ utilizes the full observation set.
The estimator $\boldsymbol{\phi}$ generates the estimated context $\mathbf{z}_t$, explicitly predicting external states.
The input to the high-level policy $\pi^H$, denoted as $\mathbf{o}_t$, includes the gravity unit vector $\mathbf{g}_t$, joint positions and velocities $\mathbf{q}_t, \dot{\mathbf{q}}_t$, the previous low-level action $\mathbf{a}_{t}$, global body yaw $\boldsymbol{\psi}_t$, ball position $\mathbf{p}_t^\text{ball}$ and user commands $\mathbf{c}^H_t$. The user commands $\mathbf{c}^H_t$ specify the target ball velocity $\mathbf{v}_t^\text{cmd}$ in the global frame.
All the above information is obtained from the quadruped robot’s proprioceptive sensors, except for the ball position.

\subsubsection{Action Space}
The high-level actor $\pi^H$ produces two outputs: an index $\mathbf{d}_t$ determines the selected low-level skill $\pi^L$ and low-level commands $\mathbf{c}^L_t$. 
Since dribbling and locomotion skills require low-level commands of different dimensions, a straightforward approach is to set the command output dimension as the sum of these individual dimensions. As skills within the same category share behavioral consistency, they can use the same portion of the low-level commands. 
Consequently, the low-level commands $\mathbf{c}^L_t$ is defined with five dimensions: the first two control dribbling skills $\pi^L_1$ and $\pi^L_2$ (corresponding to $\mathbf{v}^\text{cmd}_x$ and $\mathbf{v}^\text{cmd}_y$), while the remaining three control locomotion skills $\pi^L_3$ and $\pi^L_4$ (corresponding to $\mathbf{v}^\text{cmd}_x$, $\mathbf{v}^\text{cmd}_y$, and $\boldsymbol{\omega}$).

\vspace{0.2em}
\subsubsection{Environment Design}
We design five types of terrain—flat ground, ramp-up, ramp-down, rough terrain, and stair descent—to train the robot's ball dribbling ability in complex environments. At the start of every episode, the robot is randomly placed on one of these terrains with a random yaw orientation, and its joint angles are initialized around a nominal pose. The soccer ball is randomly positioned within a 2-meter radius of the robot. User commands $\mathbf{c}^H_t$ employ curriculum learning for progressive sampling, as explained in Sec. \ref{sec:training strategy}. Every episode lasts 20 seconds.

\section{Dynamic Skill-Focused Policy Optimization}\label{sec:DSF-PO}

For high-level policy training, the standard PPO procedure can be suboptimal due to asymmetric skill execution. When the skill index $\mathbf{d}_t$ is in $\{1,2\}$, only the first two dimensions of $\mathbf{c}^L_t$ are used for dribbling, yet \emph{all} dimensions receive gradient updates. This introduces two main issues. First, unused command dimensions (e.g., locomotion commands during dribbling) still produce gradients, interfering with optimization and slowing convergence. Second, the policy network wastes capacity to optimize these irrelevant parameters, reducing overall learning efficiency.

To tackle these problems, we propose Dynamic Skill-Focused Policy Optimization (DSF-PO), a novel surrogate loss formulation that dynamically adjusts optimization weights based on skill selection probabilities. This approach ensures that the policy network optimizes only the relevant command dimensions for the chosen skill. Below, we provide a detailed formulation and implementation of DSF-PO.

\subsection{Formulation}
Consider a policy $\pi_\theta(a|s)$ with two output heads: a discrete skill selector $\pi^d_\theta(d|s)$ and a continuous command generator $\pi^c_\theta(c|s, d)$. The skill selector $\pi^d_\theta(d|s)$ outputs a skill index $d \in \{1, \dots, K\}$, indicating which skill is selected. The command generator $\pi^c_\theta(c|s, d)$ outputs a command $c$, which is a union of different commands corresponding to each skill, $c = \bigcup_{k=1}^K c^k$, where $c^k \in \mathbb{R}^{n_k}$ is the command for skill $k$ with dimension $n_k$.
This formulation corresponds to a hierarchical learning problem: once a skill index $d$ is selected, the low-level network processes only the command $c^d$ while discarding other commands. 

Considering $d$ follows a categorical distribution:
\begin{equation}
    d \sim \text{Categorical}(p_1, \dots, p_K),
\end{equation}
where the probability of selecting skill $d$ is given by
\begin{equation}
    p_d = \pi_\theta^d(d|s) = \frac{\exp{z_d}}{\sum_{k=1}^K \exp{z_k}},
\end{equation}
where $z_d$ are the unnormalized logits for the skills.

The command for the selected skill $d$ follows a multivariate normal distribution:
\begin{equation}
    c^d \sim \mathcal{N}(\mu^d, \Sigma^d), \quad \mu^d = \pi^c_\theta(s, d).
\end{equation}
Thus, the overall policy $\pi_\theta(a|s)$ can be decomposed as:
\begin{equation}
    \pi_\theta(a|s) = \pi^d_\theta(d|s) \cdot \prod_{k=1}^K \mathcal{N}(c^k;\mu^k_\theta,\Sigma^k)^{\mathbb{I}(k=d)},
\end{equation}
where $\mathbb{I}(k=d)$ is an indicator function that selects the appropriate command $c^k$ based on the selected skill index $d$.

Let the skill focus weights be $w_k(s) = \pi^d_\theta(k|s)$, representing the probability of selecting skill $k$ given the state $s$. The original importance ratio for policy optimization in PPO $r_t(\theta) = \frac{\pi_\theta(a_t|s_t)}{\pi_{\theta_{\text{old}}}(a_t|s_t)}$ can be written as:
\begin{equation}
    r_t^{\text{DSF}}(\theta) = 
    \underbrace{\frac{\pi^d_\theta(d_t|s_t)}{\pi^d_{\theta_{\text{old}}}(d_t|s_t)}}_{\text{Skill ratio}} \cdot \prod_{k=1}^K \left(\underbrace{\frac{\mathcal{N}(c_t^k;\mu^k_\theta,\Sigma^k)}{\mathcal{N}(c_t^k;\mu^k_{\theta_{\text{old}}},\Sigma^k)}}_{\text{Command ratio}}\right)^{w_k(s_t)\cdot \mathbb{I}(k=d_t)} \text{.}
\end{equation}

The DSF-PO surrogate loss function can then be formulated as:
\begin{equation}
    \mathcal{L}^{\text{surrogate}} = \mathbb{E}_t\left[\min\left(r_t^{\text{DSF}}\hat{A}_t, \text{clip}(r_t^{\text{DSF}},1-\epsilon,1+\epsilon)\hat{A}_t\right)\right],
\end{equation}
where $\hat{A}_t$ is the estimated advantage at time $t$ and $\epsilon$ is the clipping parameter in PPO.

A partially correct but intuitive understanding of DSF-PO is that, according to the policy gradient theorem\cite{sutton1999policy}, the resulting policy gradient is as follows:
\begin{align}
\nabla_\theta \mathcal{L} \propto \sum_k w_k(s_t) & \left[  \nabla_\theta \log \pi^d_\theta(k \mid s_t) \right. \nonumber \\
\quad + &  \mathbb{I}(k = d_t)  \nabla_\theta \log \pi^c_\theta(c^k \mid s_t) \Big] \hat{A}_t.
\end{align}
The greater the tendency of the current policy to favor skill $k$, the larger the gradient weight associated with its command parameter.

\subsection{Implementation}
We employ PPO with Asymmetric Actor-Critic (AAC)\cite{pinto2017asymmetric} to train the actor $\pi^H$ of the high-level policy. The actor receives observations $\mathbf{o}_t$, while the value network (critic) gets the full state $\mathbf{s}_t$, which also includes the real velocities of the robot's base and the ball. 

As mentioned before, the action space comprises a discrete skill index and a continuous command, requiring different network architectures and activation functions. The actor's network begins with a Shared Feature Extractor (SFE)—a three-layer MLP $[512,256,128]$ with ELU activations—before branching into two output heads.
The index head $\pi^d$ consists of a linear layer and softmax to produce a normalized four-dimensional categorical distribution, from which the skill index $\mathbf{d}_t$ is sampled. The command head $\pi^c$ consists of a linear layer and tanh activation, generating a five-dimensional continuous output within $[-1,1]$. This output serves as the mean of a normal distribution $\mathcal{N}(\mu_t, \Sigma_t)$, from which the low-level commands $\mathbf{c}^L_t$ are sampled. The first two dimensions correspond to dribbling commands, and the remaining three correspond to locomotion commands.

Then we can formulate the action to be $\mathbf{a}_t = (\mathbf{d}_t,\mathbf{c}^d_t)$ where:
\begin{subequations}
\begin{align}
    \mathbf{d}_t \sim \text{Categorical}(p_1,...,p_4),& \quad p_k = \pi^d_\theta(k|\mathbf{s}_t).  \\
    \mathbf{c}^d_t \sim \mathcal{N}(\mu^d, \Sigma^d),& \quad \mu^d = \pi^c_\theta(\mathbf{s}_t,\mathbf{d}_t). 
\end{align}
\end{subequations}
For skill $k$ at state $\mathbf{s}_t$, its skill focus weight is:
\begin{equation}
    w_k(\mathbf{s}_t) = \pi^d_\theta(k|\mathbf{s}_t).
\end{equation}

Dribbling skills, including $\pi^L_1$ and $\pi^L_2$, share a common subset of commands output $\mathbf{c}_t$. A more appropriate notation is  $\mathbf{c}_t^k \in C_k$, where $C_k$ represents the set of all commands associated with skill $k$. Therefore, we take

\begin{equation}
    r_t^{\text{DSF}^\prime}(\theta) = 
   {\frac{\pi^d_\theta(\mathbf{d}_t|\mathbf{s}_t)}{\pi^d_{\theta_{old}}(\mathbf{d}_t|\mathbf{s}_t)}} \cdot \prod_{k=1}^K \left({\frac{\mathcal{N}(c_t^k;\mu^k_\theta,\Sigma^k)}{\mathcal{N}(c_t^k;\mu^k_{\theta_{old}},\Sigma^k)}}\right)^{w_k(\mathbf{s}_t)\cdot\mathbb{I}(\mathbf{c}_t^k \in C_k)}
\end{equation}
as the importance ratio of the DSF-PO.

Then, the surrogate loss becomes:
\begin{equation}
    \mathcal{L}^\text{surrogate} = \mathbb{E}_t\left[\min\left(r_t^{\text{DSF}^\prime}\hat{A}_t, \text{clip}(r_t^{\text{DSF}^\prime},1-\epsilon,1+\epsilon)\hat{A}_t\right)\right].
\end{equation}

\section{Training Strategy of the High-level Policy}\label{sec:training strategy}

We present a comprehensive training strategy inspired by \cite{Ji2023DribbleBot} to deal with the challenges of skill switching and training stability in high-level ball dribbling control.
To strike a balance in skill transitions, the high-level policy operates at a lower inference frequency (e.g., 10 Hz) than low-level skills.
Additionally, we incorporate carefully designed reward and curriculum learning tailored to terrains and commands, as detailed in the following.

\subsection{Reward Design}
As the high-level policy should minimize its focus on the robot's low-level actions, we employ as few reward terms as possible for training, which consist of five components shown in Table \ref{tab:reward_table}:
\begin{itemize}
    \item Encouraging the robot to maintain its balance.
    \item Encouraging the robot to approach and orient toward the ball.
    \item Rewarding the consistent output of the same skill index $\mathbf{d_t}$ over time.
    \item Rewarding the proximity between the ball's actual and expected velocity.
    \item Encouraging the robot to use dribbling skills when approaching the ball.
\end{itemize}

To ensure stable training, a bounded summation function is employed to aggregate all reward terms, with exponential kernels for weighted integration.

\subsection{Curriculum Learning} \label{subsec:curriculum learning}
Due to the high-level policy being trained from scratch, the use of curriculum learning significantly contributes to the stability of the training process. We implement a terrain curriculum to help the robot adapt to complex terrains and a command curriculum to improve the robot's ability to follow wide-range user commands effectively.
Terrain difficulty, serving as a metric for the terrain curriculum's difficulty, is employed as a parameter to control the steepness and ruggedness of the stair descent, ramp-up, ramp-down, and rough terrains, respectively.

We define $p_{\mathbf{c}^H, \mathbf{t}}^k$ as the joint distribution of the user commands $\mathbf{c}^H$ and terrain difficulty $\mathbf{t}$ used in the sampling process during the $k$-th episode.
And an efficient approach is to maintain independent distributions over $p_{\mathbf{c}^H}, p_{\mathbf{t}}$ such that $p_{\mathbf{c}^H, \mathbf{t}} = p_{\mathbf{c}^H}\cdot p_{\mathbf{t}}$.
Referring to \cite{margolis2024rapid}, we employ the box adaptive curriculum update rule with a reward-based approach.
Specifically, we initialize $p_{\mathbf{c}^H, \mathbf{t}}^0$ as a uniform probability distribution over $(\mathbf{c}^H \in [-0.5, 0.5], \mathbf{t} \in \{0, 1\})$.
The user commands and terrain difficulty for the robot are sampled independently at episode $k$: $\mathbf{c}^H \sim p_{\mathbf{c}^H}^k, \mathbf{t} \sim  p_{\mathbf{t}}^k$.
If the robot succeeds in this region, we will expand the sampling distribution to include neighboring regions, thereby potentially increasing the task's difficulty. 

\begin{table}[t]
\centering
\bgroup
\def\arraystretch{1.5}
\caption{Reward terms for high-level policy training.}
\label{tab:reward_table}
\scriptsize
\begin{tabular}{lrr}
\cline{1-3}
\multicolumn{3}{|c|}{\textbf{Reinforcement Learning}} \\
\cline{1-3}
{Term}            & {Expression} & {Weight}                    \\
\hline
Projected Gravity    &    $|\mathbf{g}_\mathrm{xy}|^2$ & -5.0         \\
Robot Ball Distance         & exp\{$-\delta_{p} |\mathbf{b}-\mathbf{p}_{\mathrm{FRHip}}|^2$\}  & 4.0  \\
Yaw Alignment   & exp\{$-\delta_{\psi}(e_{\mathrm{rbcmd}}^2+e_{\mathrm{rbbase}}^2)$\}  & 4.0   \\
Consistent Skill Index  &   $\sum_{i=t-T}^t \mathbb{I}(\mathbf{d}_i = \mathbf{d}_{i-1})$   &  0.1   \\
Change Skill Index     &   $\mathbb{I}(\mathbf{d}_t \neq \mathbf{d}_{t-1})$    &   -0.005    \\
Ball Velocity Norm     & exp$\{-\delta_n( |\mathbf{v}^\mathrm{cmd}|-|\mathbf{v}^b| )^2\} $        & 8.0          \\ 
Ball Velocity Angle     & 1$-\left(\psi_b-\psi_\mathrm{cmd}\right)^2/\pi^2$      & 8.0         \\
Ball Velocity  Error    & exp\{$-\delta_{v}|\mathbf{v}^b - \mathbf{v}^\mathrm{cmd}|^2$\}    & 8.0    \\
Dribbling Near Ball     & $\mathbb{I}(|\mathbf{b}-\mathbf{p}|<d_{\mathrm{max}})\cdot \mathbb{I}(\mathbf{d_t} \in \{1, 2\})$   & 1.0   \\

\cline{1-3}
\multicolumn{3}{|c|}{\textbf{Curriculum Learning}} \\
\cline{1-3}
{Term}            & {Expression}                    \\
\hline
$r_{\mathbf{c}^H}$  &  $\mathbb{I}(r_{\mathrm{BallVelocityError}}>0.5)$  \\
$r_{\mathbf{t}}$   & $\mathbb{I}(\mathbf{b} > d_\mathrm{th1}) \cdot \mathbb{I}(|\mathbf{b}-\mathbf{p}|<d_{\mathrm{th2}})$ & \\
\hline
\end{tabular}
\egroup
\vspace{-0.3cm}
\end{table}

Suppose the robot receives rewards $r_{\mathbf{c}^H}$ and $r_{\mathbf{t}}$ for attempting to follow $\mathbf{c}^H$ and traverse the terrain with difficulty $\mathbf{t}$, respectively. We then apply the update rule:
\begin{subequations}
\label{eq:curriculum update}
\begin{align}
p_{\mathbf{c}^H}^{k+1}\left(\mathbf{c}^H_{\mathrm{n}}\right) &\leftarrow 
\begin{cases}
p_{\mathbf{c}^H}^k\left(\mathbf{c}^H_{\mathrm{n}}\right) & \text{if } r_{\mathbf{c}^H}=1, \\
1 & \text{otherwise}.
\end{cases} \\
p_{\mathbf{t}}^{k+1}\left(\mathbf{t}_{\mathrm{n}}\right) &\leftarrow 
\begin{cases}
p_{\mathbf{t}}^k\left(\mathbf{t}_{\mathrm{n}}\right) & \text{if } r_{\mathbf{t}}=1, \\
1 & \text{otherwise}.
\end{cases}
\end{align}
\end{subequations}
Eq. \eqref{eq:curriculum update} indicates that the probability density on neighbors $\mathbf{c}^H_{\mathrm{n}}$ of $\mathbf{c}^H$ and $\mathbf{t}_{\mathrm{n}}$ of $\mathbf{t}$ are increased. 
In our case, $\mathbf{c}^H_{\mathrm{n}} \in \{ \mathbf{c}^H-0.1,\mathbf{c}^H+0.1 \}, \mathbf{t}_{\mathrm{n}} \in \{\mathbf{t}+1\}$.
The reward terms are listed in Table \ref{tab:reward_table}.

\section{Results}
In this section, we design experiments to evaluate the effectiveness of the proposed method and compare its ball dribbling performance on rugged terrains with previous approaches.
Velocity-tracking experiments are not conducted, as the ball’s velocity on uneven terrain is predominantly influenced by gravity and terrain effects.

\subsection{Simulation Performance} \label{subsec:simulation results}

\begin{figure}[h]
    \centering
    \begin{subfigure}{0.49\linewidth}
        \centering
        \includegraphics[width=\linewidth]{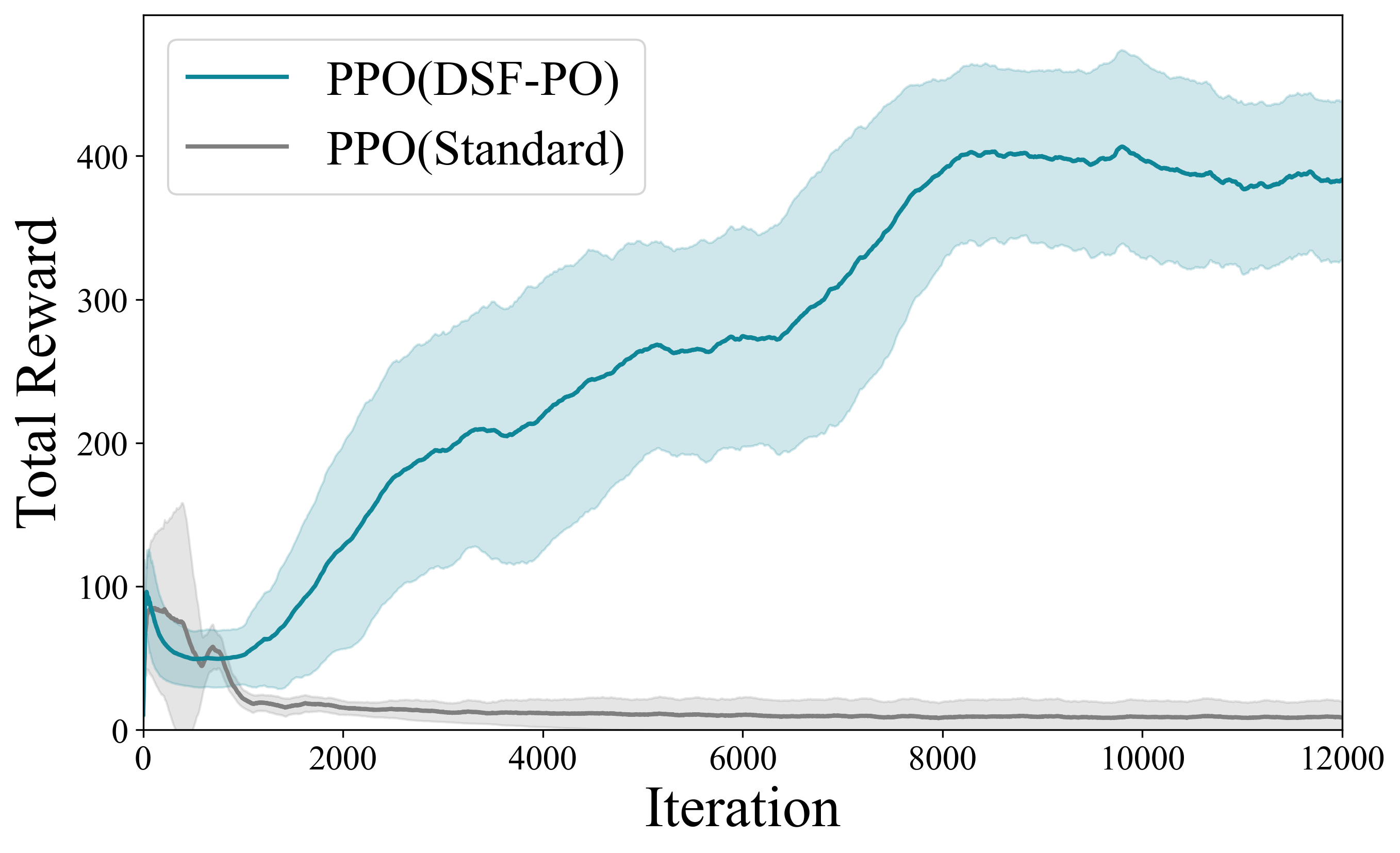}
        \caption{Average total reward}
    \end{subfigure}
    \hfill
    \begin{subfigure}{0.49\linewidth}
        \centering
        \includegraphics[width=\linewidth]{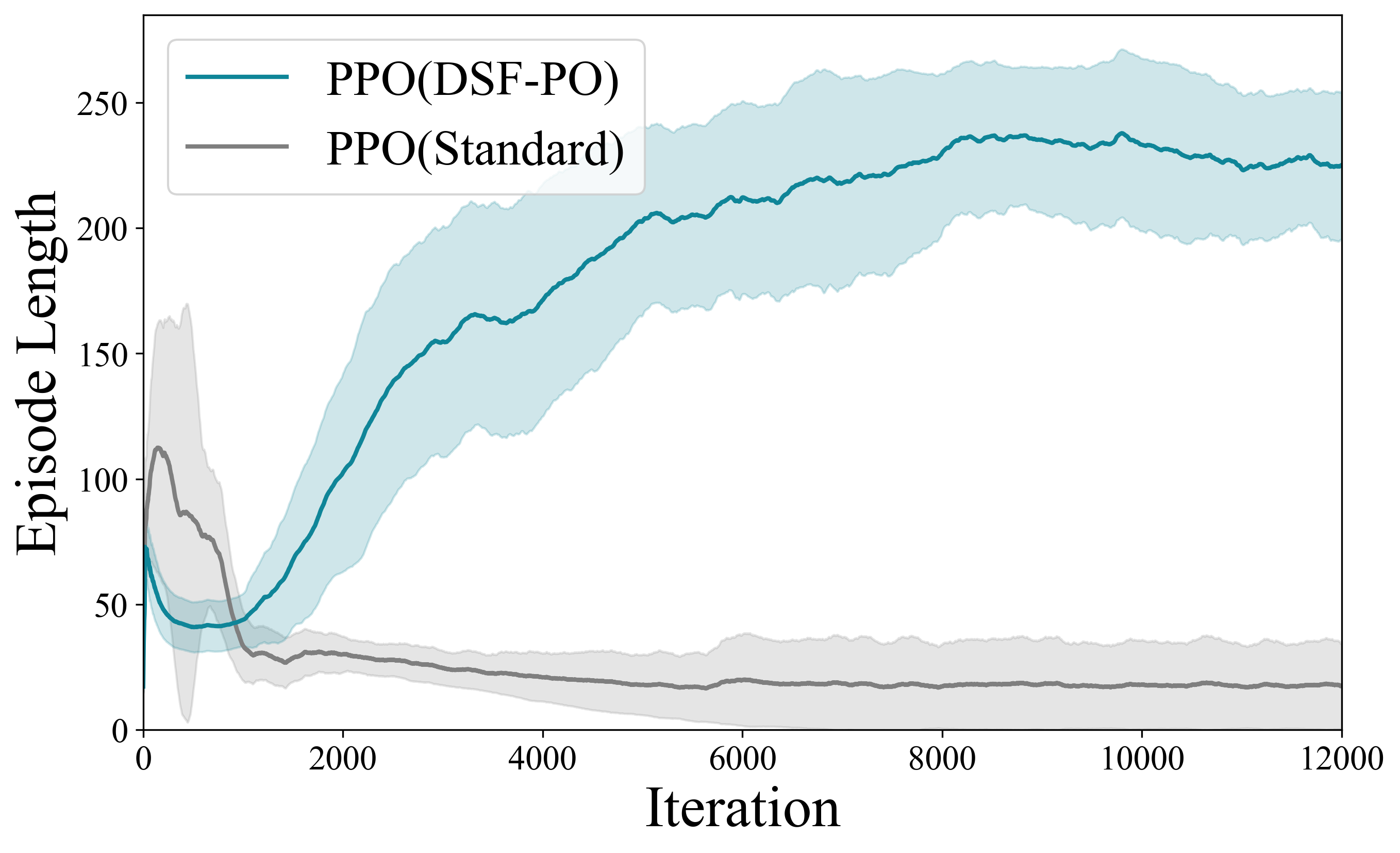}
        \caption{Average episode length}
    \end{subfigure}
    \caption{\textbf{Training curves of PPO with DSF-PO compared to standard PPO.}
    The shaded regions indicate the standard deviation over multiple runs. 
    }
    \label{fig:training curve}
    \vspace{-0.3cm}
\end{figure}

\begin{figure*}[t]
    \centering
    \includegraphics[width=1.0\textwidth, clip, trim={50 0 0 0}]{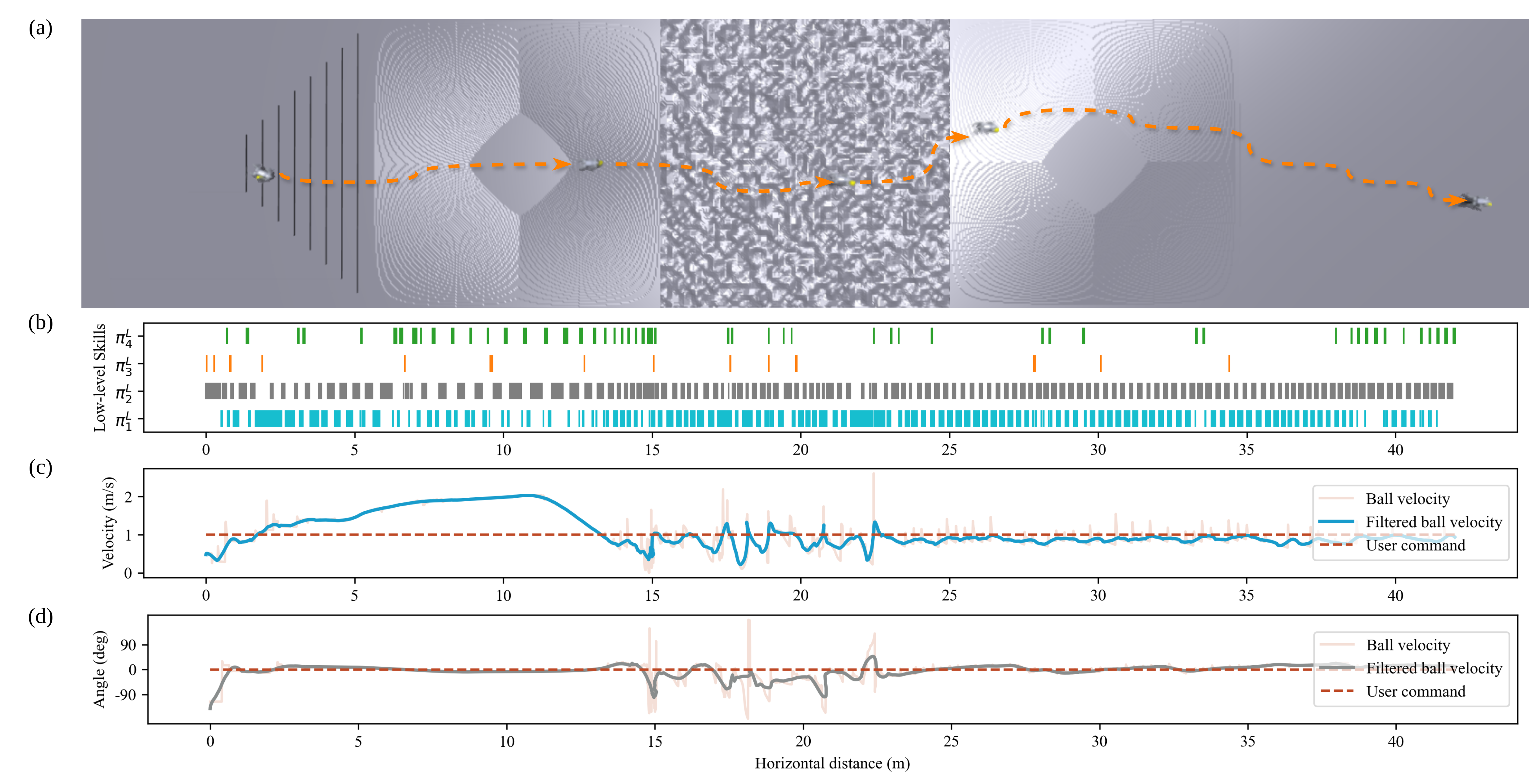}
    \caption{\textbf{Cross-terrain dribbling performance evaluation.}
    (a) A trajectory schematic of the ball dribbling across five terrains in sequence: stair descent, ramp-down, rough terrain, ramp-up, and flat ground. Each terrain measures \SI{10}{\meter} per side.
    (b) Visualization of the invocation of different low-level skills, where each thin vertical line represents a single invocation.
    (c-d) Visualization of the ball's velocity magnitude and direction.
    The horizontal axis represents the robot’s traveled horizontal distance.
    }
    \label{fig:switching}
\end{figure*}

\subsubsection{Learning Performance}
We demonstrate the learning performance of the high-level policy trained with PPO.
The simulation and training are conducted in Isaac Gym using an NVIDIA RTX 3090 Ti.
An ablation study is performed to evaluate the performance improvement of DSF-PO compared to standard PPO.
Both methods use the same hyperparameters, with the only difference being the calculation of the loss function.

Fig. \ref{fig:training curve} shows that the models are trained over \num{12000} iterations, with performance measured in total reward and episode length. PPO with DSF-PO achieves higher rewards and longer episode lengths, steadily improving throughout training, while standard PPO plateaus earlier. This suggests that DSF-PO’s adaptive optimization based on skill selection improves exploration and prevents premature convergence, enabling more robust policy learning.

\subsubsection{Terrain Traversability}
To assess the terrain traversability of the learned policy, we compare our model against DribbleBot\cite{Ji2023DribbleBot} and DexDribbler\cite{hu2024dexdribbler}. Since their publicly available checkpoints were trained on the Unitree Go1, we conduct evaluations using Go1's URDF to ensure a fair comparison. All policies are tested under the same command: $\mathbf{c}_t = (1.0, 0.0)$. 

We conduct \num{100} tests per method on each terrain, where success is defined as the robot reaching the terrain boundary. As shown in Table \ref{tab:comparison}, all methods achieve 100\% success on flat ground, but performance varies on more challenging terrains. Our approach outperforms the baselines on ramp-up (92\%), ramp-down (97\%), rough terrain (80\%), and stair descent (78\%). These results highlight our method's superior adaptability and robustness across diverse terrains.

\begin{table}[h]
\centering
\vspace{0.22cm}
\bgroup
\def\arraystretch{1.5}
{
\begin{tabular}{cclrccc}
\cline{3-5}
 & & \multicolumn{3}{|c|}{Flat Ground} & &   \\
\cline{3-5} 
 & & DB & $100\%$ & \multirow{2}{*}{\parbox{0.20\linewidth}{ \centering \vspace{0.1cm} \includegraphics[width=\linewidth,clip,trim={25 75 25 25}]{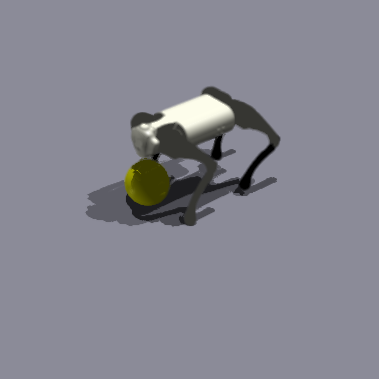}}} & &  \\
 & & DD & $100\%$ & & & \vspace{0.05cm}\\
 & & Ours & $100\%$ & & & \vspace{0.05cm} \\
\end{tabular}

\begin{tabular}{lrcclrc}
\cline{1-3}
\cline{5-7}
\multicolumn{3}{|c|}{Ramp-Up} & & \multicolumn{3}{|c|}{Ramp-Down}  \\
\cline{1-3}
\cline{5-7}
DB & $89\%$ & \multirow{3}{*}{\parbox{0.20\linewidth}{ \centering \vspace{0.1cm} \includegraphics[width=\linewidth,clip,trim={50 90 50 90}]{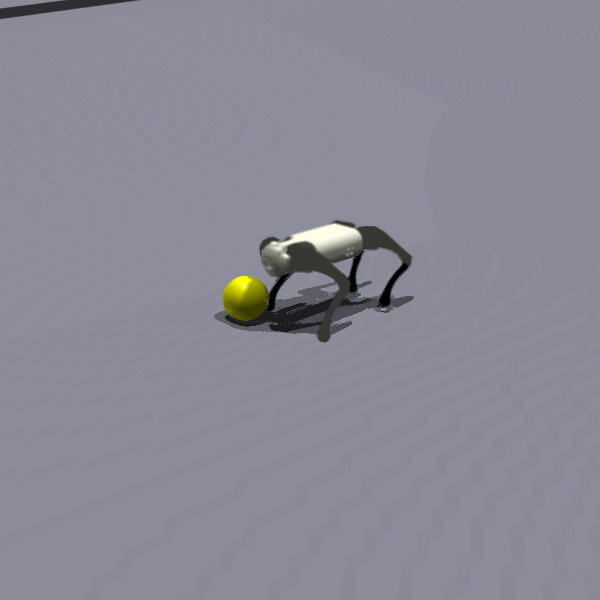}}} &  & DB & $90\%$ \  & \multirow{3}{*}{\parbox{0.20\linewidth}{ \centering \vspace{0.1cm} \includegraphics[width=\linewidth,clip,trim={50 90 50 90}]{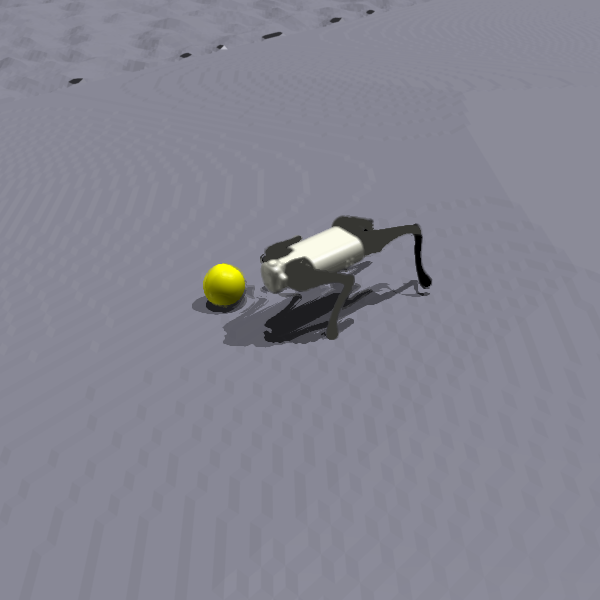}}} \\
DD & $91\%$ & & & DD & $92\%$ \vspace{0.05cm}\\
\textbf{Ours} & $\mathbf{92\%}$ & & & \textbf{Ours} & $\mathbf{97\%}$ \vspace{0.05cm} \\
 
\cline{1-3}
\cline{5-7}
\multicolumn{3}{|c|}{Rough Terrain} & & \multicolumn{3}{|c|}{Stair Descent}  \\
\cline{1-3}
\cline{5-7}
DB & $53\%$ & \multirow{3}{*}{\parbox{0.20\linewidth}{ \centering \vspace{0.1cm} \includegraphics[width=\linewidth,clip,trim={80 110 75 105}]{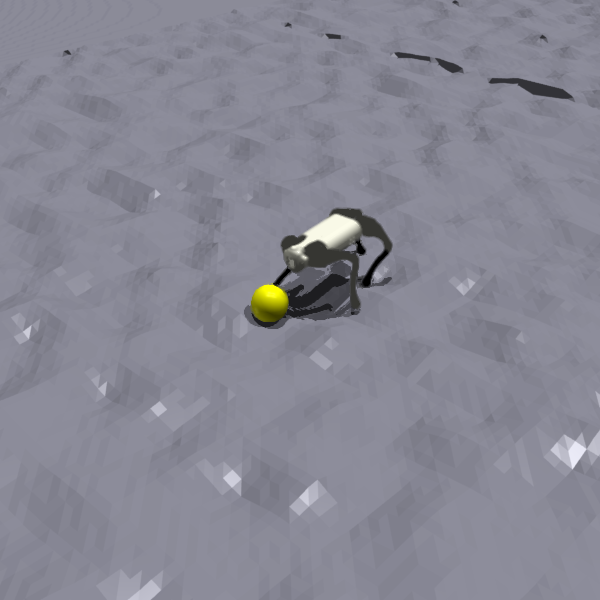}}} &  & DB & $61\%$ \  & \multirow{3}{*}{\parbox{0.20\linewidth}{ \centering \vspace{0.1cm} \includegraphics[width=\linewidth,clip,trim={50 90 50 90}]{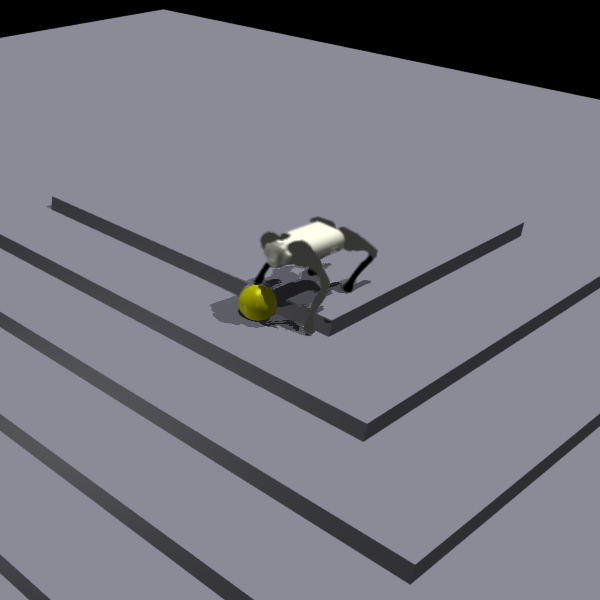}}} \\
DD & $22\%$ & & & DD & $13\%$ \vspace{0.05cm} \\
\textbf{Ours} & $\mathbf{80\%}$ & & & \textbf{Ours} & $\mathbf{78\%}$ \vspace{0.05cm} \\
\hline
\end{tabular}
}
\egroup
\caption{\textbf{Comparison of ball dribbling success rates across different terrains.}
The success rates of three methods—DribbleBot (DB), DexDribbler (DD), and our approach—are evaluated under various terrains: flat ground, ramp-up (\textit{slope=\num{0.1}}), ramp-down (\textit{slope=\num{-0.1}}), rough terrain (\textit{elevation\_difference=\SI{0.1}{\meter}}), and stair descent (\textit{height=\SI{0.05}{\meter}, width=\SI{0.5}{\meter}}).
}
\label{tab:comparison}
\vspace{-0.7cm}
\end{table}

\subsubsection{Cross-terrain Dribbling}

As shown in Fig. \ref{fig:switching}, we evaluate the robot’s dribbling ability across five terrains: stair descent, ramp-down, rough terrain, ramp-up, and flat ground. With a fixed command $\mathbf{c}^H_t=(0.0, -1.0)$, the robot successfully completes the task in \num{121} seconds. It maintains a stable direction except on rough terrain, where external disturbances cause deviations.

\begin{figure}[t]
    \centering
    \includegraphics[width=0.8\linewidth]{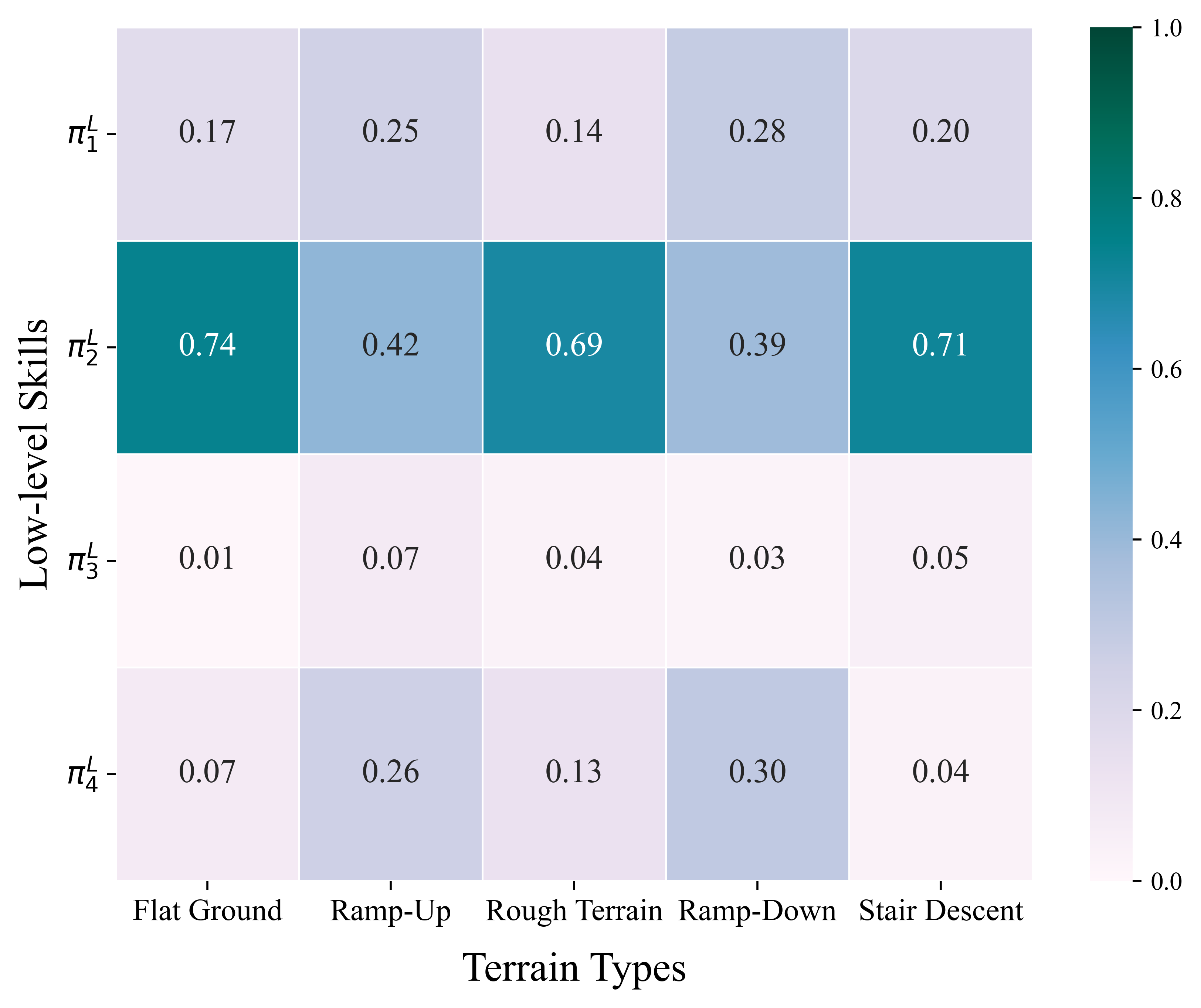}
    \caption{\textbf{Usage frequency of low-level skills across different terrains.}
    The numbers represent the proportion of each low-level skill's usage frequency on a given terrain.
    }
    \label{fig:heatmap}
    \vspace{-0.5cm}
\end{figure}

Fig. \ref{fig:switching}(b) shows the invocation of low-level skills by the high-level policy across different terrains. The results indicate that dribbling skills are used most frequently, allowing the policy to adaptively switch between them based on terrain variations. Locomotion skills are invoked less frequently, mainly when the robot needs to reposition itself after drifting away from the ball.
To systematically analyze low-level skill usage variations across terrains, we collect \num{10000} steps of data per terrain and record the skill indices selected by the high-level policy. The statistical distribution is shown in Fig. \ref{fig:heatmap}, revealing distinct behavior patterns:

\begin{itemize}
    \item Flat ground: The policy predominantly utilizes $\pi_2^L$, as it matches the training environment of this skill, ensuring stable dribbling.  
    \item Stair descent: Locomotion skills are used more often to help the robot navigate discontinuities.  
    \item Ramp-up \& Ramp-down: The skill distributions are nearly identical, suggesting a similar control strategy for both terrain types.  
    \item Rough terrain: The policy employs $\pi_4^L$ more frequently to adapt to the uneven surface and maintain stability.  
\end{itemize}  

Fig. \ref{fig:switching}(c) and \ref{fig:switching}(d) illustrate the ball’s velocity magnitude and direction, respectively. On downhill terrains (stair descent, ramp-down), the robot struggles to precisely control the ball’s velocity due to gravitational acceleration. Velocity fluctuations are pronounced on rough terrain, mainly due to uneven surfaces. In contrast, on ramp-up and flat ground, the robot effectively regulates the ball’s velocity, demonstrating greater stability and control.  

\begin{figure}[t]
    \centering
    \includegraphics[width=1.0\linewidth]{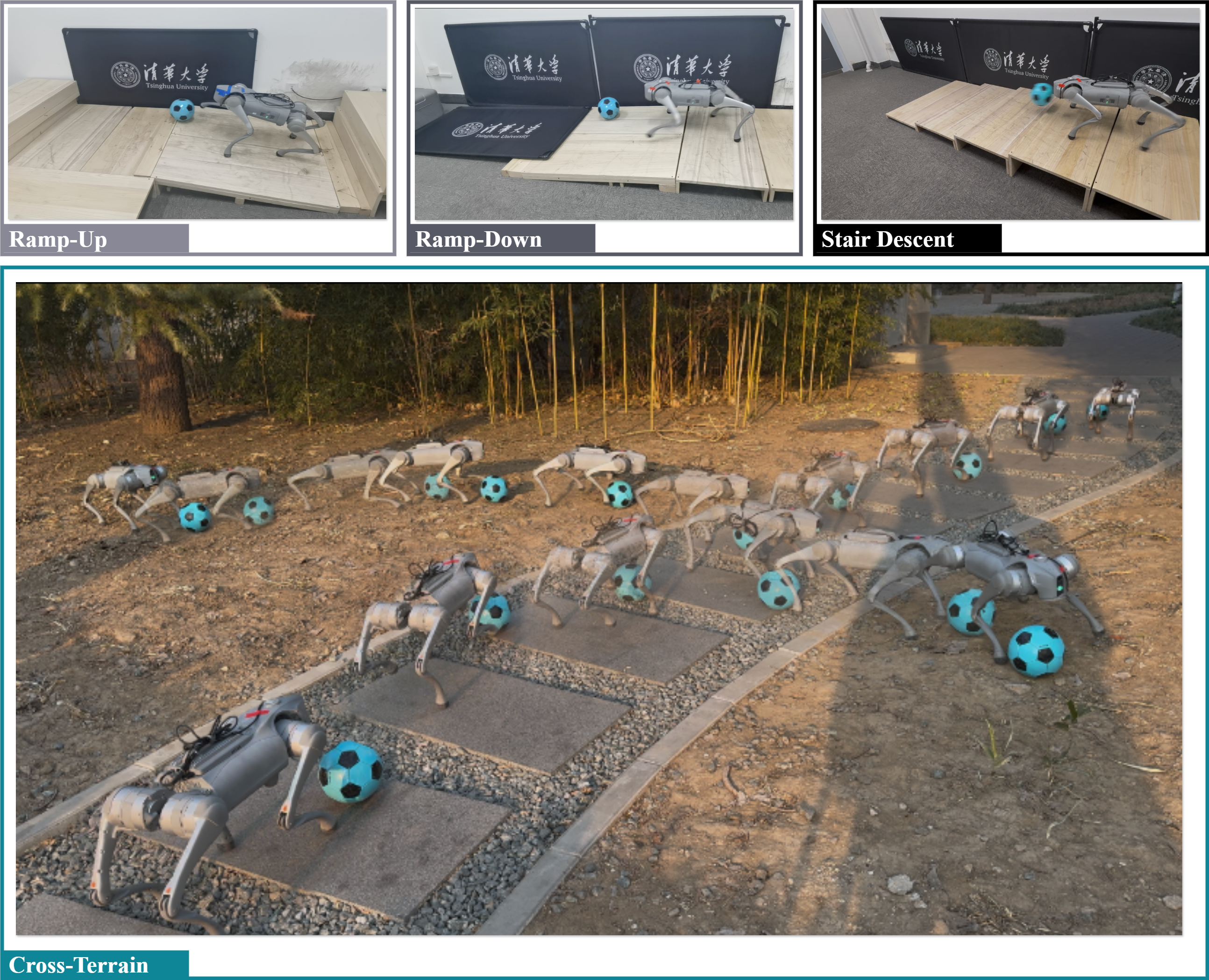}
    \caption{\textbf{Real-world deployment experiments on different terrains.}
    The indoor environment includes ramp-up, ramp-down, and stair descent, while the outdoor environment consists of a complex multi-terrain field.
    }
    \label{fig:exp}
\end{figure}

\subsection{Physical Deployment} \label{subsec:experiments}
We use a Unitree Go2 quadruped robot, which is additionally equipped with a downward-facing fisheye camera featuring a 240° field-of-view mounted on its head. All policy inference runs onboard with an NVIDIA Jetson Orin NX. 
A YOLOv11\cite{yolo11_ultralytics} network is trained to detect the ball for estimating its spatial coordinates relative to the robot. Ball localization is improved by integrating data from the front and downward-facing cameras. The fisheye equidistant model converts angles to distances, and two models compute the ball's 2D coordinates: one uses the ball’s diameter, and the other relies on the camera’s distance and angle. A Kalman filter fuses these estimates for enhanced localization.
We use the PD controller with $k_p=20, k_d=0.5$ for policy training and evaluation, while the locomotion skill $\pi^L_4$ uses different PD gain $k_p=40, k_d=1.0$ following  the original work.
All low-level and high-level policies are transferred to the physical robot in a zero-shot manner.

\begin{table}[t]
\centering
\caption{Real world ball dribbling performance evaluation.}
\label{tab:success rate}
\begin{threeparttable}
\begin{tabular}{
>{\centering\arraybackslash}p{1.5cm}|>{\centering\arraybackslash}p{1.2cm}>{\centering\arraybackslash}p{1.2cm}>{\centering\arraybackslash}p{1.2cm}>{\centering\arraybackslash}p{1.2cm}
} 
\toprule
\multirow{2}{*}{Method}& \multirow{2}{*}{Ramp-Up} & Ramp-Down& \multirow{2}{*}{Gravel} & Stair Descent \\ 
\midrule
DribbleBot* &  $0/4$ & $0/4$ & $\mathbf{4/4}$   & $-$  \\
DexDribbler* &  $\mathbf{4/5}$ & $1/5$ & $\mathbf{5/5}$  & $-$  \\
Ours &        $\mathbf{4/5}$ & $\mathbf{3/5}$ & $\mathbf{5/5}$  & $\mathbf{3/5}$   \\
\bottomrule
\end{tabular}
\begin{tablenotes}
\item[*] The data are sourced from \cite{Ji2023DribbleBot, hu2024dexdribbler}.
\end{tablenotes}
\end{threeparttable}
\vspace{-0.5cm}
\end{table}

We evaluate our policy on four terrains, as illustrated in Fig. \ref{fig:exp}.
Among them, the ramp-up, ramp-down, and stair descent terrains are constructed indoors. In the stair descent terrain, each step measures \SI{50}{\centi\meter} in width and \SI{5}{\centi\meter} in height. The cross-terrain scenario consists of irregular ground surfaces, raised curbs, smooth stone pavement, and soft gravel terrain.
Both the robot and the ball are initialized on a flat surface, with the objective of successfully moving the ball to the other end of the terrain while ensuring the robot maintains balance. High-level commands are published via a remote controller.

We conduct five trials on each terrain, and the statistical results are presented in Table \ref{tab:success rate}. Hardware limitations make it infeasible to reproduce DribbleBot and DexDribbler's performance on these terrains fairly, so we refer to their reported experimental data for comparison instead.
The results demonstrate that our method consistently achieves the highest success rate across all four terrains, highlighting its superior generalization capability. 
A detailed analysis of the trials reveals that our high-level policy enables the robot to dynamically adapt its locomotion and dribbling skills based on terrain variations. 
On flat surfaces, the robot maintains a steady dribbling gait with minimal adjustments. When navigating ramps, it adjusts limb coordination and body posture to counteract inclination effects while stabilizing ball control. During stair descents, the robot first kicks the ball down the steps before carefully following, adjusting its gait to maintain stability and control. On rough outdoor terrain, the robot actively modulates stride length and stance width, leveraging its agile skill-switching capability to overcome obstacles and sustain stable dribbling.
These observations show that the learned policy enables successful ball manipulation while closely matching simulation behaviors, demonstrating the effectiveness of our hierarchical RL framework in real-world deployment.

\section{Conclusion and Future Works}
In this paper, we propose a hierarchical RL framework that trains a high-level policy to coordinate dribbling and locomotion skills for dynamic ball manipulation on rugged terrains. We also introduce a dynamic skill-focused loss formulation to improve learning efficiency and convergence in mixed discrete-continuous action spaces.
Simulation and real-world experiments show that our methods outperform previous works in generalizing across diverse terrains, achieving stable ball dribbling performance on rugged surfaces. 
There are several directions for future works that we aim to explore and enhance: 
(1) Incorporating additional optimization techniques to accelerate DSF-PO convergence and improve sample efficiency.
(2) Extending the ball position input of low-level skills to three dimensions, mitigate positional ambiguity on uneven terrain.
(3) Integrating additional multimodal low-level skills through our hierarchical RL framework to enhance its overall autonomous behavior capabilities.
Our overarching goal is to endow the quadruped robot with advanced agility, enabling it to more effectively assist humans.

{
\bibliographystyle{IEEEtran}
\bibliography{main}
}

\end{document}